\documentclass[review]{elsarticle}
\usepackage{hyperref}
\usepackage{amsmath}
\usepackage{caption}
\usepackage{multirow}
\usepackage{color}

\def\ie{\emph{i.e. }}

\journal{Journal of \LaTeX\ Templates}

\begin{document}

\begin{frontmatter}

%\title{Elsevier \LaTeX\ template\tnoteref{mytitlenote}}
%\title{MDFN: Multi-Scale Deep Feature Learning Network for Object Detection\tnoteref{mytitlenote}}
\title{MDFN: Multi-Scale Deep Feature Learning Network for Object Detection}

\author{Wenchi Ma\fnref{myfootnote1}}
%\ead{wenchima@ku.edu}
\author{Yuanwei Wu\fnref{myfootnote1}}
%\ead{y262w558@ku.edu}
\author{Feng Cen\fnref{myfootnote2}}
%\ead{feng.cen@tongji.edu.cn}
\author{Guanghui Wang\fnref{myfootnote1}}
%\ead{ghwang@ku.edu}
%\address{The University of Kansas, Lawrence, KS, 66045}
\fntext[myfootnote1]{W. Ma, Y. Wu, and G. Wang are with the Department of Electrical Engineering and Computer Science, The University of Kansas, Lawrence, KS, 66045 USA e-mail: \{wenchima, y262w558, ghwang\}@ku.edu.}
\fntext[myfootnote2]{F. Cen is with the Department of Control Science and Engineering, College of Electronics and Information Engineering, Tongji University, Shanghai 201804, China Email: feng.cen@tongji.edu.cn}

%\address[add1]{Department of Electrical Engineering and Computer Science, The University of Kansas, Lawrence, KS, 66045 USA}
%\address[add2]{Department of Control Science and Engineering, College of Electronics and Information Engineering, Tongji University, Shanghai 201804, China}



\begin{abstract}
This paper proposes an innovative object detector by leveraging deep features learned in high-level layers. Compared with features produced in earlier layers, the deep features are better at expressing semantic and contextual information. The proposed deep feature learning scheme shifts the focus from concrete features with details to abstract ones with semantic information. It considers not only individual objects and local contexts but also their relationships by building a multi-scale deep feature learning network (MDFN). MDFN efficiently detects the objects by introducing information square and cubic inception modules into the high-level layers, which employs parameter-sharing to enhance the computational efficiency. MDFN provides a multi-scale object detector by integrating multi-box, multi-scale and multi-level technologies. Although MDFN employs a simple framework with a relatively small base network (VGG-16), it achieves better or competitive detection results than those with a macro hierarchical structure that is either very deep or very wide for stronger ability of feature extraction. The proposed technique is evaluated extensively on KITTI, PASCAL VOC, and COCO datasets, which achieves the best results on KITTI and leading performance on PASCAL VOC and COCO. This study reveals that deep features provide prominent semantic information and a variety of contextual contents, which contribute to its superior performance in detecting small or occluded objects. In addition, the MDFN model is computationally efficient, making a good trade-off between the accuracy and speed. 
\end{abstract}

\begin{keyword}
%\texttt{elsarticle.cls}\sep \LaTeX\sep Elsevier \sep template
deep feature learning\sep multi-scale \sep semantic and contextual information \sep small and occluded objects.
%\MSC[2010] 00-01\sep  99-00
\end{keyword}

\end{frontmatter}

%\linenumbers

\section{Introduction}

Recent development of the convolutional neural networks (CNNs) has brought significant progress in computer vision, pattern recognition, and multimedia information processing \cite{xu2019toward,8794588,xu2019adversarially, yang2015scene}. As an important problem in computer vision, object detection has a lot of potential applications, such as image retrieval, video surveillance, intelligent medical, and unmanned vehicle~\cite{he2018learning,mo2018efficient}. In general, the progress is mainly contributed by the powerful feature extraction ability of CNN through building deeper or wider hierarchical structure given the great advancement of computer hardware. The ideas of deep and residual connections~\cite{he2016deep, huang2016deep}, network-in-network and inception structures~\cite{szegedy2015going}, multi-box and multi-scale techniques~\cite{liu2016ssd} and dense blocks~\cite{huang2017densely} consistently dedicate to enhancing the feature expression by extracting more effective features information, especially the details from shallow layers, and to maximize the transmission of various information between the two ends of the network. Features produced in earlier layers attract more attention due to the fact that they hold more original information and details. This is beneficial to hard detection tasks like small object detection~\cite{wu2019unsupervised}. However, effective features have the possibility of being changed, attenuated or merged through the long process of forward transmission in deep and complicated networks~\cite{li2018detnet}. On the other hand, efficient object detection depends not only on detail features, but also on semantic and contextual information, which is able to describe both the relationships among different objects and the correlation between objects and their contexts~\cite{chen20182}. Furthermore, by processing feature information through the complicated hierarchical structure and transmitting it across multiple layers would reduce the efficiency of feature learning and increase the computational load~\cite{wang2018non}. 

\begin{figure}[h]
	\centering
	\includegraphics[width=0.7\linewidth]{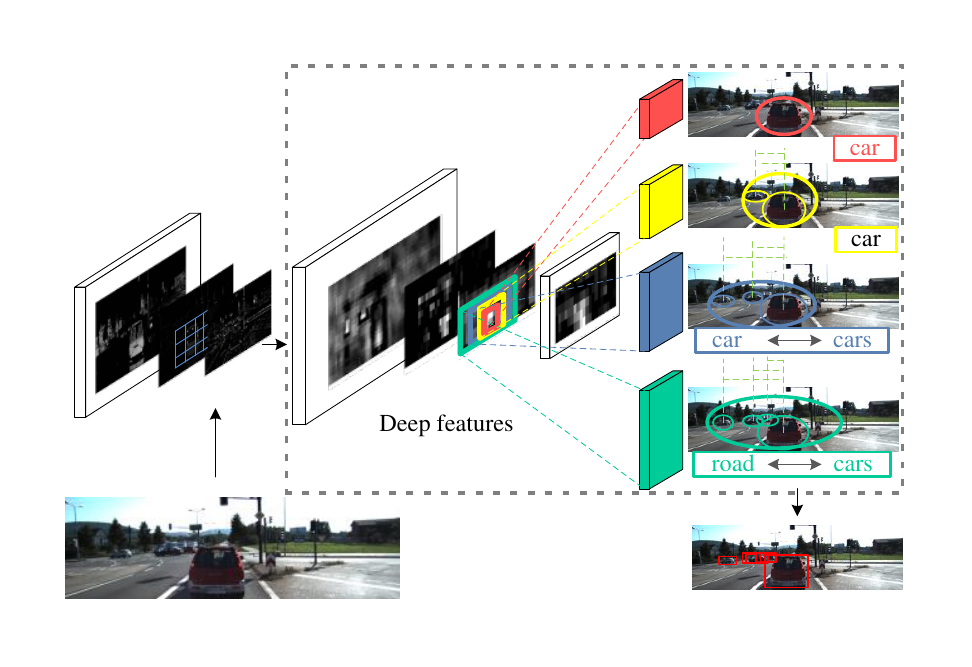}\\%
	%(a)\hspace{0.42\linewidth}(b)\\%[2pt]
	\caption{Motivation of multi-scale deep feature learning network (MDFN) for object detection}
	\vspace{-2mm}
	\label{fig:detectionrecog1}
\end{figure}

In this study, we intend to alleviate the above problem by making full use of deep features produced in the deep part of a network. We propose a multi-scale deep feature learning network (MDFN) which will generate the abstract and semantic features from the concrete and detailed features yield at the top of the base networks. It integrates the contextual information into the learning process through a single-shot framework. As shown in Figure~\ref{fig:detectionrecog1}, the semantic and contextual information of the objects would be activated by multi-scale receptive fields on the deep feature maps. The red, yellow, blue and green components represent four sizes of filters, which correspond to different object expressions. For example, the red one tends to be sensitive only to the red vehicle in the middle, while the yellow and the blue ones may also cover the small cars around it, due to the semantic expression towards correlations among different object cars. The green one has the largest activation range, and it not only detects all vehicles but also the road, by utilizing the semantic description of the relationships between the objects and their background. This process to extract various semantic information can be realized in deep layers where the receptive fields are able to cover larger scenes and feature maps produced in deep layers, which already own the abstract ability of semantic expression~\cite{chen20182, li2018detnet}. We find that most available classical networks are powerful enough in feature extraction and are able to provide necessary detail features. Motivated by these observations, we adopt the transfer learning model and design an efficient multi-scale feature extraction units in the deep layers that are close to the top of the network. The extracted deep feature information is fed directly to the prediction layer. We propose four inception modules and incept them in four consecutive deep layers, which are used to extract the contextual information. These modules significantly extend the ability of various feature expression, from which a deep feature learning based multi-scale object detector is realized.

The feature maps produced in deep layers are believed to be used for removing irrelevant contents and extracting semantic and the most important characteristics of objects against background. In contrast, the activation towards feature maps from earlier layers are supposed to extract various details, such as textures or contours of objects or their background details~\cite{mahendran2015understanding}. Currently, most deep convolutional neural networks suffer from the detection of small and occluded objects, which has not been well solved even with much more complicated models~\cite{liu2016ssd}. In our study, we claim that the detection of small and occluded objects depends not only on detail features but also on semantic features and the contextual information~\cite{wang2018context}. Deep features have better expression towards the main characteristics of objects and more accurate semantic description of the objects in the scenes~\cite{li2018detnet, wang2018non}. MDFN can effectively learn the deep features and yield compelling results on popular benchmark datasets.

From the perspective of overall performance, feature maps produced in earlier layers have higher resolutions than those produced in the deep part, which brings about the difference of computational load. Thus, the increase of filtering operation in the latter ones would not introduce heavy computational burden, especially when compared to the frameworks with very deep or wide base networks where a great amount of filters are designed in the shallow layers employs a relatively small base network, VGG-16, which is neither too deep nor too wide. Moreover, MDFN further decreases the model size by constructing \emph{information square and cubic inception modules} which are able to efficiently share parameters of filters. In addition, we extract multi-scale feature information by feeding feature outputs from different levels of layers directly to the final output layer. This strengthens the propagation of feature information and shortens the paths between two ends of the network so as to enhance the efficiency of feature usage and make the model easier to train~\cite{huang2017densely}.

The main contributions of this study include:

\begin{itemize}
\item We propose a new model that focuses on learning the deep features produced in the latter part of the network. It makes full use of the semantic and contextual information expressed by deep features. Through the more powerful description of the objects and the surrounding background, the proposed model provides more accurate detection, especially for small and occluded objects;
\item We propose the deep feature learning inception modules, which are able to simultaneously activate multi-scale receptive fields within a much wider range at a single layer level. This enables them to describe objects against various scenes. Multi-scale filters are introduced in these modules to realize the proposed information square and cubic convolutional operation on the deep feature maps, which increases the computational efficiency by parameter sharing at the same time;
\item We investigate how the depth of the deep feature learning affects the object detection performance and provide quantitative experimental results in terms of average precision (AP) under multiple Intersection over Union (IoU) thresholds. These results show substantial evidence that features produced in the deeper part of networks have a prevailing impact on the accuracy of object detection.
\end{itemize}

In addition, the proposed MDFN models outperform the state-of-the-art models on KITTI~\cite{geiger2012we}, and achieves a good trade-off between the detection accuracy and the efficiency on PASCAL VOC 2007~\cite{everingham2007pascal} and COCO~\cite{lin2014microsoft} benchmarks. MDFN has a better portability as a relatively small network module. The proposed multi-scale deep feature extraction units can be easily incepted into other networks and be utilized for other vision problems. The proposed model and source code can be downloaded from the author's website.

\section{Related Work}

\paragraph{Feature Extraction:} As a fundamental step of many vision and multimedia process tasks, feature extraction and representation has been widely studied~\cite{gao2016novel, mukherjee2015comparative, szegedy2017inception}, especially at the level of network structures, that attracted a lot of attention in the deep learning field. Deeper or wider networks amplify the differences among architectures and gives full play to improve feature extraction ability in many computer vision applications~\cite{cen2019dictionary}. The skip-connection technique~\cite{he2016deep} solved the problem of gradient vanishing to certain degree by propagating information across layers of different levels of the network, shortening their connections, which stimulates the hot research in constructing much deeper networks and have obtained improved performance. From the advent of LeNet5~\cite{lecun1998gradient} with 5 layers to VGGNet with 16 layers~\cite{russakovsky2015imagenet}, to ResNet~\cite{he2016deep} which can reach over 1000 layers, the depth of networks has dramatically increased. ResNet-101~\cite{he2016deep} shows its advantage of feature extraction and representation, especially when being used as base network for object detection tasks. Many researchers tried to replace the base network with ResNet-101. SSD~\cite{liu2016ssd,fu2017dssd} achieved its better performance with Residual-101 on PASCAL VOC2007~\cite{everingham2007pascal}. RRC~\cite{ren2017accurate} adopted ResNet as its pre-trained base network and yielded competitive detection accuracy with the proposed recurrent rolling convolutional architecture. However, SSD only obtained 1\% of improvement for mAP~\cite{fu2017dssd} by replacing the VGG-16 with the Residual-101, while its detection speed decreases from 19 FPS to 6.6 FPS, which is almost three times' drop. VGG network won in the second place in ImageNet Large Scale Visual Recognition Challenge(ILSVRC) 2014. It is shallow and thin with only 16 layers, which is another widely-used base network. Its advantage lies in the provision with a trade-off between the accuracy and the running speed. SSD achieved its best general performance by combining VGG-16 as the feature extractor with the proposed multi-box object detector in an end-to-end network structure.

Another approach to enhancing the feature extraction ability is to increase the network width. GoogleNet~\cite{szegedy2015going} has realized the activation of multi-size receptive fields by introducing the inception module, which outputs the concatenation of feature-maps produced by filters of different sizes. GoogleNet ranked the first in ILSVRC 2014. It provided a feature expression scheme of inner layer, which has been widely adopted in later works. The residual-inception and its variances~\cite{szegedy2017inception, szegedy2016rethinking} showed their advantage in error-rate over individual inception and residual technique. SqueezeDet~\cite{wu2017squeezedet} achieved state-of-the-art object detection accuracy on the KITTI validation dataset with a very small and energy efficient model, which is based on inner-layer inception modules and continuous inter-layer bottleneck filtering units in the year of 2017.

\paragraph{Attention to Deep Features:} Stochastic depth based ResNet improves the training for deep CNN by dropping layers randomly, which highlights that there is a large amount of redundancy in propagation process~\cite{huang2016deep}. The research of Viet \textit{et al}. proved by experiments that most gradients in ResNet-101 come only from 10 to 34 layers' depth~\cite{veit2016residual}. On the other hand, a number of methods draw multi-scale information from different shallow layers based on the argument that small object detection relies on the detail information produced in earlier layers. While experiments show that semantic features and objects' context also contribute to the small object detection, as well as for occlusion~\cite{chen20182}. DSSD adopts deconvolution layers and skip connections to inject additional context, which increases feature map resolution before learning region proposals and pool features~\cite{fu2017dssd}. Mask R-CNN adds mask output extracted from much finer spatial layout of the object. It is addressed by the pixel-to-pixel correspondence provided by small feature maps produced by deep convolutions~\cite{he2017mask}.

\paragraph{SSD:} The single-shot multi-box detector (SSD) is one of the state-of-the-art object detectors through its multi-box, multi-scale algorithm~\cite{liu2016ssd}. It is generally composed of a base network (feature extractor) and a feature classification and localization network (feature detector). The base network, VGG-16, is pre-trained on ImageNet and then being transferred to the target dataset through transfer learning. SSD discretizes its output space of bounding boxes into a set of default boxes over various aspects ratios and scales for each feature map location and it realizes classification and localization by regression with the multi-scale feature information from continuous extraction units~\cite{liu2016ssd}. SSD has the advantage over other detectors for its trade-off between higher detection accuracy and its real-time detection efficiency.

\section{Deep Feature Learning Network}

\begin{figure*}[h]
	\centering
	\includegraphics[width=1.0\linewidth]{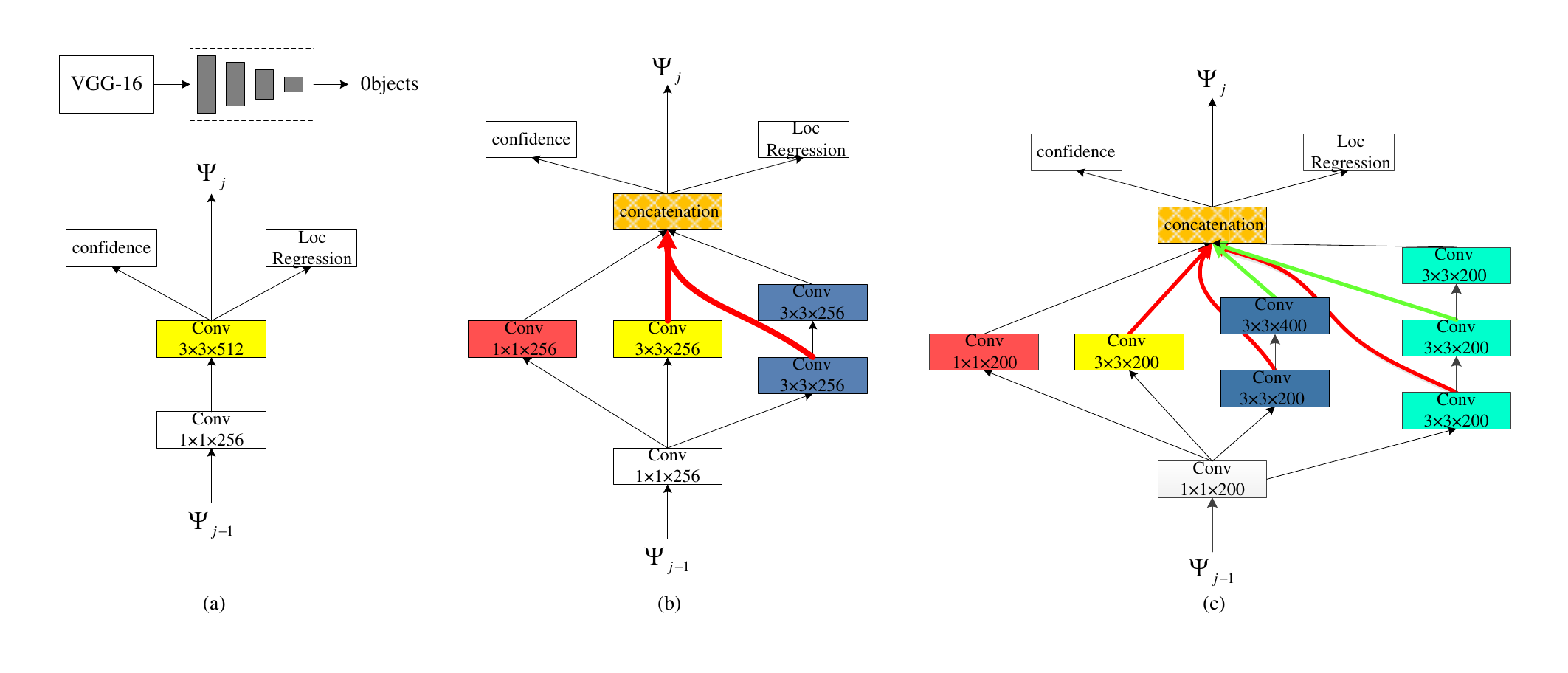}\\%
	%(a)\hspace{0.42\linewidth}(b)\\%[2pt]
	\caption{\textbf{Deep feature learning inception modules.} (a) The core and basic deep feature transmission layer structure. (b) and (c) denote the two actual layer structures of information square and cubic inception modules. Red and green arrows indicate the way of parameter sharing. $\Psi_{j-1}$ represents feature maps from previous layer and $\Psi_{j}$ denotes the output feature maps from current layer.}
	\label{fig:detectionrecog3}
\end{figure*}

MDFN aims at efficiently extracting deep features by constructing deep feature learning inception modules in the top four layers, as shown in Figure~\ref{fig:detectionrecog2}. MDFN builds a multi-scale feature representation framework by combining multi-box, multi-scale and multi-level techniques. The MDFN is conceptually straightforward: a single-shot object detection model with a pre-trained base network, which scales a small size and maintains a high detection efficiency. The overall structure of MDFN is shown in Figure~\ref{fig:detectionrecog2}, and the detailed analysis of each module is given below. 

\subsection{Deep Feature Extraction and Analysis}

\paragraph{Deep Features:} With the increase of layers, feature maps produced from the deep part of the network become abstract and sparse, with less irrelevant contents, serving for the extraction of the main-body characteristics of the objects~\cite{mahendran2015understanding}. These feature maps have smaller scales but correspond to larger receptive fields. This determines their function of deep abstraction to distinguish various objects with better robust abstraction so that features are relatively invariant to occlusion~\cite{ren2017accurate}. Since the feature maps from intermediate levels retrieve contextual information either from their shallower counterparts or from their deeper counterparts~\cite{ren2017accurate}, the extraction of deep features from consecutive layers is necessary. On one hand, our proposed deep feature learning inception modules directly process high-resolution feature information from base network and their outputs are directly fed to the output layers, which shortens the path of feature propagation and enhances the efficiency of feature usage. On the other hand, multi-scale filtering on deep feature maps further strengthens the extraction of the semantic information. As a result, the multi-scale learning for deep features increases the dimension of the extracted information and enhances the ability of semantic expression.

\paragraph{Deep Multi-scale Feature Extraction:} Feature extraction in CNN can be expressed as a series of non-linear filtering operations as follows~\cite{ren2017accurate}.

\begin{equation}
\Psi_{n}=f_{n}(\Psi_{n-1})=f_{n}(f_{n-1}(...f_{1}(X))) 
\end{equation}
\noindent where $\Psi_{n}$ refers to the feature map in layer $n$, and $f_{n}$ denotes the $n$-th nonlinear unit which transforms the feature map from layer $n-1$ to the $n$ layer.
\begin{equation}
%\begin{aligned}
\hat{O}=O(T_{n}(\Psi_{n}),...,T_{n-l}(\Psi_{n-l})),n>l>0
%\end{aligned} 
\end{equation}

In equation (2), $T_{n}$ is the operation transmitting the output feature maps from the $n$-th layer to the final prediction layer. Thus, equation (2) is an operation of multi-scale output; and $O$ stands for the final operation that considers all input feature maps and then provides final detection results. 

According to \cite{ren2017accurate}, equation (2) performs well relying on the strong assumption that each feature map being fed into the final layer has to be sufficiently sophisticated to be helpful for detection and accurate localization of the objects. This is based on the following assumptions: 1) These feature maps should be able to provide the fine details especially for those from the earlier layers; 2) the function that transforms feature maps should be extended to the layers that are deep enough so that the high-level abstract information of the objects can be built into the feature maps; and 3) the feature maps should contain appropriate contextual information such that the occluded objects, small objects, blurred or overlapping ones can be inferred exactly and localized robustly~\cite{ren2017accurate, stewart2016end, li2018detnet}. Therefore, the features from both the shallow and deep layers play indispensable roles for the object recognition and localization. Moreover, the feature maps from the intermediate levels retrieve contextual information either from their shallower counterparts or from their deeper counterparts~\cite{ren2017accurate}. Thus, some work tries to make the full utilization of the features throughout the entire network and realizes connections across layers as many as possible, so as to maximize the probability of information fusion including both the details and contexts, like DenseNet~\cite{huang2017densely}. 

Although maximizing information flow across most parts of the network would be able to make full use of feature information, it also increases the computational load. Most importantly, there is a possibility that this intense connection across layers may not achieve the expected effectiveness as the negative information would also be accumulated and passed during the transmission process, especially in the deep layers~\cite{wang2018non}. Furthermore, features with low intensity values are easy to be merged~\cite{liu2015parsenet}. The above analysis can be shown in the following equation.

\begin{equation}
\Psi_{n}+\delta_{n}=f_{n}(\Psi_{n-1}+\delta_{n-1})=f_{n}(f_{n-1}(...f_{n-p}(\Psi_{n-p}+\delta_{n-p}))),n>p>0 
\end{equation}
where $\delta_{n}$ is the accumulated redundancy and noise that existed in layer $n$, and $\delta_{n-p}$ is the corresponding one accumulated in shallow layers.

Based on the above analysis, in order to efficiently exploit the detected feature information, another constrained condition should be considered so as to prevent the features from being changed or overridden. We claim that the feature transmission across layers should decrease the probability of features being changed by drift errors or overridden by the irrelevant contents, and should minimize the accumulation of the redundancy and noise especially in the deep layers. Thus, feature transmission within the local part of the network or direct feature-output should be a better solution to effectively use this information. To this end, we propose the following multi-scale deep feature extraction and learning scheme, which will support the above strong assumptions and satisfy the related conditions.

\begin{equation}
\Psi_{m}=F_{m}(\Psi_{m-1})=F_{m}(F_{m-1}(...F_{m-k}(\Psi_{n}))), m-k>n 
\end{equation}
\begin{equation}
\begin{aligned}
F_{j}=S(&f_{j}(f_{j}(f_{j}(\Psi_{j-1}))),f_{j}(f_{j}(\Psi_{j-1})),\\
&f_{j}(\Psi_{j-1}), \Psi_{j-1};W_{j}), m-k \leq j \leq m
\end{aligned} 
\end{equation}
\begin{equation}
\hat{O}=O(T_{m}(\Psi_{m}),T_{m-1}(\Psi_{m-1}),...,T_{m-k}(\Psi_{m-k}),T_{m-j}(\Psi_{m-j})),j>k
\end{equation}

\noindent where $m$ indicates high-level layers, $\Psi_{m}$ is the corresponding output feature maps of layer $m$. The function $F_{j}$ maps $\Psi_{j-1}$ to multi-scale spatial responses in the same layer $j$. $F$ is functioned by S, the feature transformation function, and weighted by $W$. All feature information produced in high-level layers would be directly fed to the final detection layer by the function $T$. $\Psi_{m-j}$ represents the feature map from some shallow layer. The inputs of function $O$ include feature maps from low-level layers and those from high-level layers.

The above functions (4), (5), and (6) construct a deep multi-scale feature learning scheme. It considers feature maps produced from shallow layers which have high resolution to represent fine details of objects. This is in accordance with the first mentioned assumption. $F_{m}$ are designed for deep layers of the network to introduce deep abstraction into the output feature maps. Moreover, multi-scale receptive fields within a single deep layer are sensitive to most important features and objects in the contexts of different sizes, which makes the output powerful enough to support the detection and localization, and directly responds to the strong suggestion. At the same time, several continuous deep inception units provide the probability that feature maps from intermediate levels can retrieve contextual information from both lower and deeper counterparts. This is beneficial to detecting the exact locations of overlapped objects, occluded, small, and even blurred or saturated ones which need to be inferred robustly~\cite{stewart2016end}. This satisfies the above assumptions 2) and 3).

Instead of building connections across layers, the consecutive deep inception realizes the same function of multi-scale feature maps, abstraction and contextual information built-in and simultaneously avoids the problem of introducing redundancy and noise as described in the proposed constrained condition. Moreover, the multi-scale inceptions would produce more variety of information, rather than simply increase the information flow by connections across the layers. Based on the above analysis, the proposed model makes the training smoother and achieves a better performance of localization and classification.

\subsection{Deep Feature Learning Inception Modules}

Deep feature learning inception modules capture the direct outputs from the base network. Our basic inception module makes full use of the deep feature maps by activating multi-scale receptive fields. In each module, we directly utilize the output feature information from the immediate previous layer by $1\times1$ filtering. Then, we conduct 3$\times$3, 5$\times$5 and 7$\times$7 filtering to activate various receptive fields on the feature maps so as to capture different scopes of the scenes on the corresponding input images. We realize the multi-scale filtering only with the 1$\times$1 and 3$\times$3 filters in practice to minimize the number of parameters~\cite{szegedy2016rethinking, ma2018mdcn}. We build two types of power operation inception modules for the high-level layers: one is information square inception module, and the other is information cubic inception module, as shown in Figure~\ref{fig:detectionrecog3}. We build these two modules by assigning weights to different filters as given in the following equations, where the two operations are denoted by $F_{j}^{2}$ and $G_{j}^{3}$, respectively.

\begin{equation}
F_{j}^{2}(\Psi_{j-1})=f_{j}(f_{j}(\Psi_{j-1}))+2 \times f_{j}(\Psi_{j-1})+\Psi_{j-1}, m-k \leq j \leq m 
\end{equation}
\begin{equation}
\begin{aligned}
G_{j}^{3}(\Psi_{j-1})=&g_{j}(g_{j}(g_{j}(\Psi_{j-1})))+3 \times g_{j}(g_{j}(\Psi_{j-1}))+\\
&3 \times g_{j}(\Psi_{j-1})+\Psi_{j-1}, m-k \leq j \leq m
\end{aligned} 
\end{equation}

\noindent where the 5$\times$5 filer is replaced by two cascaded 3$\times$3 filters and the 7$\times$7 filter is replaced by three cascaded 3$\times$3 filters. This replacement operation has been verified to be efficient in~\cite{szegedy2016rethinking}. The number of parameters of the two cascaded 3$\times$3 filters only accounts for 18/25 of that of one single 5$\times$5 filter~\cite{szegedy2016rethinking}. By manipulation, the expressions of (7) and (8) can actually be approximated by the following information square and cubic operations, respectively. 

\begin{equation}
% F_{j}=(f_{j}(\Psi_{j-1}))+\Psi_{j-1})^{2}, m-k \leq j \leq m
F_{j}^{2}(\Psi_{j-1}) = (f_{j}^{2}+2 \times f_{j} +1)(\Psi_{j-1}) = ((f_{j}+1)^{2})(\Psi_{j-1}) 
\end{equation}

\begin{equation}
%F_{j}=(f_{j}(\Psi_{j-1}))+\Psi_{j-1})^{3}, m-k \leq j \leq m
G_{j}^{3}(\Psi_{j-1}) =(g_{j}^{3}+3 \times g_{j}^{2}+3 \times g_{j} +1)(\Psi_{j-1}) = ((g_{j}+1)^{3})(\Psi_{j-1})
\end{equation}

\paragraph{Parameter Sharing:} The proposed information square and cubic inception modules can be implemented efficiently by sharing parameters. For example, we share parameters between the 3$\times$3 and 5$\times$5 filtering units by extracting outputs from the first 3$\times$3 filter of the 5$\times$5 unit and concatenate it with the parallel outputs from the 3$\times$3 filtering unit. Then, the number of output tunnels of the 3$\times$3 filtering operation is implicitly doubled while the set of filters are only used once, as indicated by the red arrows in Figure~\ref{fig:detectionrecog3} (b). This parameter-sharing can be further used in the cubic inception module as shown in Figure~\ref{fig:detectionrecog3} (c). The outputs of the 3$\times$3 filtering operation come from the 3$\times$3, 5$\times$5, and 7$\times$7 filtering unit respectively as indicated by the three red arrows in Figure~\ref{fig:detectionrecog3} (c). Similarly, those outputs of the 5$\times$5 filtering operation come from the 5$\times$5 and the 7$\times$7 filtering unit respectively as shown by the two green arrows.

\begin{figure*}[h]
	\centering
	\includegraphics[width=1.1\linewidth]{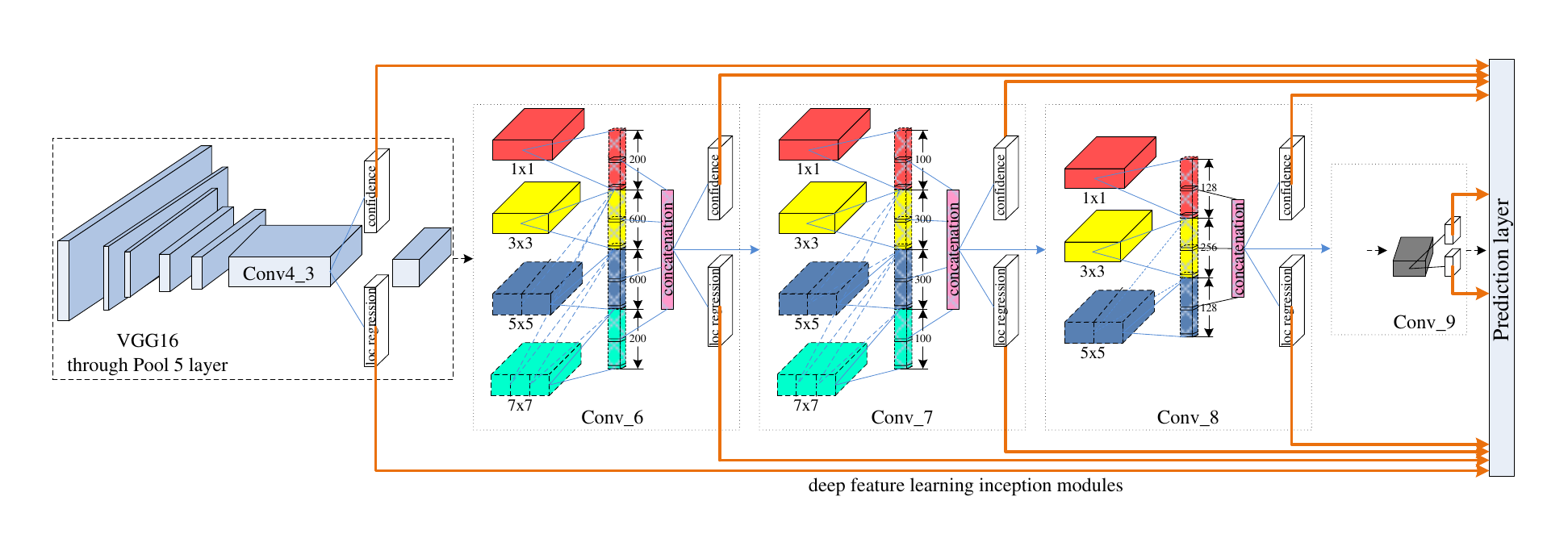}\\%
	%(a)\hspace{0.42\linewidth}(b)\\%[2pt]
	\caption{\textbf{The architecture of MDFN.} The proposed deep feature learning inception modules are introduced in the layers of Conv\_6, Conv\_7, Conv\_8 (and Conv\_9 in MDFN\_I2). Each one employs filters of multiple sizes, which are represented by the red, yellow, blue and green boxes. The white boxes refer to classification and localization regression layers for the jointly learning of multi-scale detection~\cite{liu2016ssd}. }
	\label{fig:detectionrecog2}
\end{figure*}

\subsection{Multi-Scale Object Detection Scheme}

Single shot multi-box detector (SSD)~\cite{liu2016ssd} is a popular and effective object detection model. One of the key techniques proposed in SSD is multi-box matching and sifting, which discretizes the output space of bounding boxes into a group of default boxes with various aspect ratios and scales per feature map location~\cite{liu2016ssd}. In the test process, SSD produces scores for the presence of every object category in each default box and generates several choices of the bounding boxes to better match the object shape. Specifically, the offsets relative to the default boxes and the scores of every classes, in each feature map cell, indicating the existence of object class instance in each box are predicted. Inspired by SSD, we propose a multi-scale object detection scheme. In our model, given $k$ boxes to each given location, calculates $c$ class scores and four offsets of the four vertexes of each box relative to the default box. This ends up with a total of $k(c+4)$ filters serving for each location inside the feature map. Thus, the number of outputs should be $k(c+4)mn$ for each feature map with the dimension of $m \times n$. It has been verified in~\cite{liu2016ssd} that using various default box shapes would facilitate the task of predicting boxes for the single-shot network, which increases the accuracy of object localization and classification. We adopt this multi-box technique as the first property of our multi-scale scheme.

In addition, we extract feature maps with various resolutions from multi-level outputs. MDFN combines feature maps with different resolutions from both shallow and deep layers of the network, as shown in Figure~\ref{fig:detectionrecog2}. Our deep feature learning inception modules are applied in the four consecutive high-level layer units. These four layer units transmit their output deep features directly to the final prediction layer, which shortens the information transmission to the full extent. From the perspective of training, these shortened connections make the input and output of the network closer to each other, which benefits the training of the model, even though the deep inceptions make high-level layers complicated~\cite{huang2017densely}. The direct connections between the high-level layers and the final prediction layer alleviate the problem of vanishing gradient and strengthen the feature propagation~\cite{huang2017densely}. On the other hand, the sequence of four high-level layer units maximizes the ability of deep feature extraction and representation by the semantic and contextual information acquisition in two ways. First, it makes the latter three high-level layers obtain the contextual information from the previous lower layers. Second, layers in the same level can provide the contextual information of different scopes and more exact semantic expressions that would be naturally built in the current layer outputs. This process is the second property of our multi-scale scheme.

We use the multi-scale filters to active the receptive fields of various sizes, which aims to enhance the extraction of the semantic and contextual information. Another aspect to note is the sizes of feature maps. In most networks, sizes of the feature maps would be gradually reduced with the increase of depth. This considers the limited memory of our systems and the scale invariance of the features. Thus, our multi-scale filters that are incepted in the deep part of the network would have less computational burden as the resolutions of their input feature maps are much smaller than those produced in the earlier layers. This counteracts the computational load brought by the increase of filtering operations. This is the third property of the proposed multi-scale scheme.

The introduction of this multi-scale scheme maximizes the acquisition of multi-scale feature information including the deep abstraction and various contextual-related contents. This supports the strong suggestion of multi-scale feature application mentioned above. The deep feature learning network makes full use of the abstract but highly semantic feature information and solves the balance of increasing computational load and the operation efficiency of the network. We will show this advantage by quantified data results in the experiments section.

\paragraph{Base Network:} MDFN utilizes the VGG-16~\cite{russakovsky2015imagenet} as its base network for feature extraction, and adopts transfer learning as the basic learning style for object detection. Thus, we employ the VGG-16 network pre-trained on ImageNet~\cite{krizhevsky2012imagenet} dataset and then make our high-level network transferred to the target datasets. VGG-16 has only 16 convolutional layers if only the convolution and pooling layers are taken into consideration. This VGG network is constructed only by the 3$\times$3 filter, which is regarded as one of the most efficient filter size~\cite{iandola2016squeezenet}. Following SSD, the prediction layer of the MDFN combines the low-level features from the layer of $conv4\_3$ at the depth of 13, of which output feature maps are with the resolution of 38$\times$38. This provides direct high-resolution features along with sufficient details for hard object detection tasks. VGG-16 is not the best network in terms of the performance in feature extraction, however, it is widely used since it makes a good trade-off between the detection accuracy and the test speed. %Here shows the correlation between network top-5 error and number of layers from several common networks in Figure 5 (where fully-connected layers are taken into consideration in this figure, so there are 19 layers for VGG network). 
The details of the connection between VGG-16 and our proposed deep feature learning framework are shown in Figure~\ref{fig:detectionrecog2}. In order to validate the usefulness and generalization ability of the proposed MDFN modules, we also build network model with ResNet-101 pre-trained on ImageNet ~\cite{krizhevsky2012imagenet} as the base network and use it as a reference for the PASCAL VOC experiment.

\paragraph{Layer Structure of Deep Inception Module:} We propose two deep feature learning network architectures, referred as MDFN-I1 and MDFN-I2 respectively. They both have four high-level deep feature inception units. Each unit is composed of a 1$\times$1 convolution layer and the following multi-scale filtering layer, where the 1$\times$1 convolution is introduced as the bottleneck layer to reduce the number of input feature-map channels~\cite{parviainen2010dimension}. As the feature maps' sizes still keep decreasing in the deep part of the network, we introduce the proposed information cubic inception modules into the first two layers, layer $conv\_6$ and layer $conv\_7$, in both the two MDFN models, of which the output feature maps are with the resolutions of 19$\times$19, 10$\times$10 respectively. Thus, considering the input feature maps' sizes of the next two layers are already relatively small, we introduce the information square module both into the latter two high-level layers, layer $conv\_8$ and layer $conv\_9$ in MDFN-I2 and only introduce information square module into layer $conv\_8$ in MDFN-I1. We will compare the performances of the two models and analyze the impact on the deep feature learning ability. The specific configuration of inception modules are shown in Table~\ref{table:detectionrecog1}.

\begin{table*}[h]
	\centering
	\includegraphics[width=1.0\linewidth]{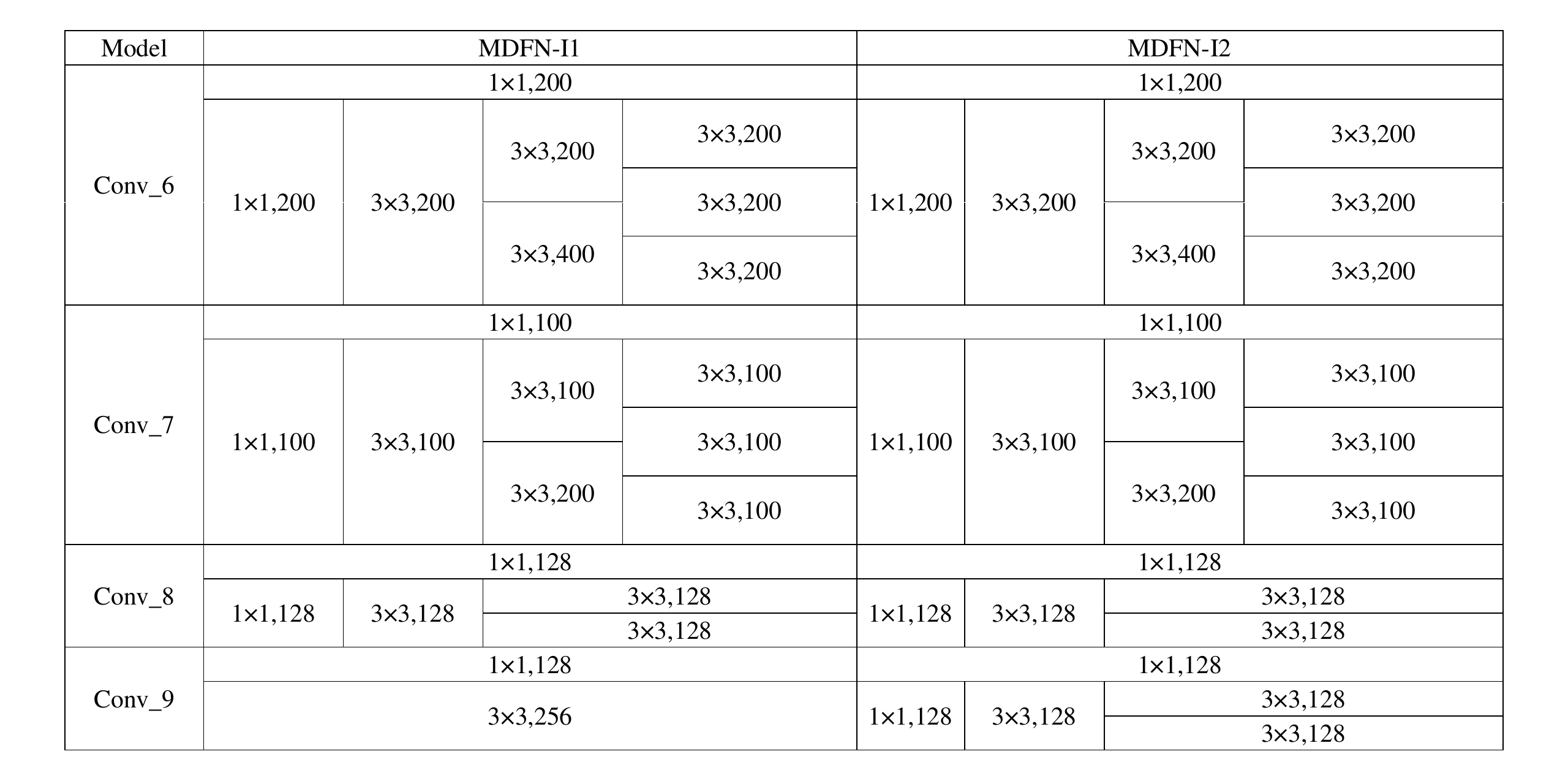}\\%
	%(a)\hspace{0.42\linewidth}(b)\\%[2pt]
	%\captionsetup[table]{skip=0pt}
	\caption{Layer structure of deep inception module layout in MDFN-I1 and MDFN-I2}
	\label{table:detectionrecog1}
\end{table*}

\paragraph{Trainig:} We employ the VGG-16 pre-trained on ImageNet as the base network of our proposed MDFN models. Then the models are fine-tuned on target dataset: KITTI or PASCAL VOC2007 or COCO.

The training was conducted on a computing cluster environment. The proposed MDFN employs the stochastic gradient descent (SGD) algorithm for training~\cite{bottou2004stochastic, zhang2018bpgrad}. Due to the limit of GPU memory, our models are trained with the mini-batch size of 16 on KITTI and 32 on PASCAL VOC and COCO. The momentum is set to 0.9, and the weight decay is 0.0005 for all the datasets, which are identical to the training approach of Liu \textit{et al}.~\cite{liu2016ssd}. The overall number of training iterations is set to 120,000 for KITTI and PASCAL VOC and 400,000 for COCO. We maintain a constant learning rate decay factor, which multiplies the current learning rate by 0.1 at 80,000 and 100,000 iterations for KITTI and PASCAL VOC, and at 280,000, 360,000 for COCO. MDFN-I1 and MDFN-I2 models adopt the learning rate of 0.0006, 0.0007 respectively on KITTI and COCO, and 0.0008, 0.0007 respectively on Pascal VOC2007. 

MDFN matches the default boxes to any ground truth box with the Jaccard overlap higher than the threshold of 0.5. MDFN imposes the set of aspect ratios for default boxes as \{1,2,3,1/2,1/3\}. We minimize the joint localization loss by smooth L1 loss~\cite{girshick2015fast} and confidence loss by Softmax loss, shown as below.

\begin{equation}
\begin{aligned}
L(x,c,l,g) = \frac{1}{N}(L_{conf}+\alpha L_{loc})
\end{aligned} 
\end{equation} 

\noindent where $N$ refers to the number of matched default boxes and the weight term $\alpha$ is set to 1~\cite{liu2016ssd}. $L_{conf}$ and $L_{loc}$ stand for the confidence loss and localization loss, respectively. 
For data augmentation, we adopt the same method as the original SSD model~\cite{liu2016ssd}. We do not use the recent random expansion augmentation trick used by the latest SSD related frameworks~\cite{fu2017dssd,liu2016ssd}.

\section{Experiments}

We empirically demonstrate the effectiveness of MDFN on the prevailing KITTI~\cite{geiger2012we}, PASCAL VOC~\cite{everingham2007pascal} and Microsoft COCO~\cite{lin2014microsoft} benchmarks. We analyze the object detection accuracy in terms of average precision (AP), and object detection efficiency in terms of speed and model sizes. We also perform a thorough comparison between the MDFN and the state-of-the-art models on these datasets. The proposed framework is implemented using Caffe~\cite{jia2014caffe}, compiled with the cuDNN~\cite{chetlur2014cudnn} computational kernels. The test speed is obtained under Titan XP GPU.

\subsection{Dataset}
\paragraph{KITTI:} KITTI object detection dataset is designed for autonomous driving, which contains challenging objects like small and occluded cars, pedestrians and cyclists. It is obtained in highway, rural and urban driving scenarios by stereo cameras and LiDAR scanners. KITTI for object detection contains 7,481 images for training and validation, and 7,518 images for testing, providing around 40,000 object labels classified as easy, moderate, and hard ones based on how much objects are occluded and truncated. Since the ground truth of the test set is not publicly available, we follow the way in~\cite{xiang2017subcategory, wu2017squeezedet}, randomly splitting the 7,381 training and validation images evenly into a training and a validation set. We evaluate the proposed MDFN models on the validation set and report the average precision (AP) on it at the three difficulty levels following the suggestion in ~\cite{geiger2012we,xiang2017subcategory}. For KITTI experiments, we scale all the input images to 1242$\times$375 and use the batch size of 16. Our models are trained to detect 3 categories of objects, including car (merged with motors), pedestrian, and cyclist. The thresholds for car, pedestrian, and cyclist are 70\%, 50\% and 50\%, respectively. All methods shown in Table~\ref{table:detectionrecog2} are obtained with the same rules described above.

\paragraph{PASCAL VOC 2007:} In VOC experiments, we follow the normal practice in the literature, the models are trained on the union of PASCAL VOC 2007 and 2012 trainval set (16,551 images) and tested on PASCAL VOC 2007 test set (4,952 images). We scale all the input images to 500$\times$500. Our models are trained to detect 20 categories of objects on VOC. The overlap threshold for each category in VOC is set to 0.5. All the methods listed in Table~\ref{table:detectionrecog6} follow the same rules as above except for the scale of input images.

\paragraph{COCO:} Microsoft COCO~\cite{lin2014microsoft} is a widely used visual recognition dataset focusing on full scene understanding. Objects in COCO contains multifarious scales and occlusion situations, where objects are smaller scaled than PASCAL VOC. We utilize the trainval35k~\cite{bell2016inside} for training and follow the strategy in~\cite{liu2016ssd}. Experiments are implemented on two image scales, 300$\times$300 and 512$\times$512. The results are shown on COCO test-dev2015 and are evaluated based on the COCO-style average precision (AP).

\subsection{Detection Results on KITTI}

\begin{table*}[ht]
	\begin{center}
		\def\arraystretch{1.0}
		\scalebox{0.6}[0.6]{%
			\begin{tabular} {|c|c|c|c|c|c|c|c|c|c|c|}
				\hline
				\multirow{2}{*}{\textbf{Model}} 
				& \multicolumn{3}{c|}{\textbf{Car}} & \multicolumn{3}{c|}{\textbf{Pedestrian}} & \multicolumn{3}{c|}{\textbf{Cyclist}} & \multirow{2}{*}{\textbf{mAP}} \\
				\cline{2-10}
				& \textbf{Easy} & \textbf{Moderate} & \textbf{Hard} & \textbf{Easy} & \textbf{Moderate} & \textbf{Hard} & \textbf{Easy} & \textbf{Moderate} & \textbf{Hard} &\\
				\hline\hline
				RPN & 82.91 & 77.83 & 66.25 & 83.31 & 68.39 & 62.56 & 56.36 & 46.36 & 42.77 & -\\
				\hline
				SubCNN & \textbf{95.77} & 86.64 & 74.07 & \textbf{86.43} & 69.95 & 64.03 & 74.92 & 59.13 & 55.03 & -\\
				\hline
				MS-CNN & 90.0 & 89.0 & 76.1 & 83.9 & 73.7 & 68.3 & 84.1 & 75.5 & 66.1 & 78.5\\
				\hline
				PNET & 81.8 & 83.6 & 74.2 & 77.2 & 64.7 & 60.4 & 74.3 & 58.6 & 51.7 & 69.6\\
				\hline
				Pie & 89.4 & \textbf{89.2} & 74.2 & 84.9 & 73.2 & 67.6 & 84.6 & 76.3 & 67.6 & 78.6\\
				\hline
				SqueezeDet & 90.2 & 84.7 & 73.9 & 77.1 & 68.3 & 65.8 & 82.9 & 75.4 & 72.1 & 76.7\\
				\hline
				SqueezeDet+ & 90.4 & 87.1 & 78.9 & 81.4 & 71.3 & 68.5 & \textbf{87.6} & 80.3 & 78.1 & 80.4\\
				\hline
				VGG16 + ConvDet & 93.5 & 88.1 & 79.2 & 77.9 & 69.1 & 65.1 & 85.2 & 78.4 & 75.2 & 79.1\\
				\hline
				ResNet50 + ConvDet & 92.9 & 87.9 & 79.4 & 67.3 & 61.6 & 55.6 & 85.0 & 78.5 & 76.6 & 76.1\\
				\hline
				SSD & 86.6 & 86.0 & 80.5 & 75.7 & 71.8 & 69.3 &  83.7 & 83.0 & 77.1 & 81.6\\
				\hline
				MDFN-I1 (ours) & 88.5 & 87.7 & \textbf{80.7} & 77.2 & \underline{74.6} & \underline{72.4} & 86.5 & \textbf{86.2} & \textbf{83.5} & \textbf{83.9} \\
				\hline
				MDFN-I2 (ours) & 87.9 & 87.1 & \underline{80.5}& 77.5 & \textbf{74.7} & \textbf{73.0} & 86.0 & \underline{85.8} & \underline{79.5} & \underline{83.8} \\
				\hline
			\end{tabular} %
		}
		\def\arraystretch{1.0}	
		\caption{Average precision(\%) on KITTI validation set. The best and second best results are highlighted in bold-face and underline fonts, respectively. }
		%\vspace{-5mm}
		\label{table:detectionrecog2}
	\end{center}
\end{table*}

\paragraph{Average Precision:} Table~\ref{table:detectionrecog2} shows the average precision (AP) of object detection on KITTI from the proposed frameworks as well as the 10 state-of-the-art models, including RPN~\cite{girshick2015fast,ren2015faster}, SubCNN~\cite{xiang2017subcategory}, MS-CNN~\cite{cai2016unified}, PNET~\cite{wu2017squeezedet}, Pie~\cite{wu2017squeezedet}, SqueezeDet~\cite{wu2017squeezedet}, SqueezeDet+~\cite{wu2017squeezedet}, VGG16 + ConvDet~\cite{wu2017squeezedet}, ResNet50 + ConvDet~\cite{wu2017squeezedet} and SSD~\cite{liu2016ssd}. From Table~\ref{table:detectionrecog2}, the proposed MDFN-I1 and MDFN-I2 networks obtain significant improvements in terms of AP, especially for the detection of pedestrian and cyclist. Its mean average precision (mAP) of all the three difficulty levels in the three categories outperforms all state-of-the-art approaches by a large margin. MDFN-I1 obtained the mAP nearly 4\% higher than that of SqueezeDet+ (a fully convolutional neural network) and almost 5\% over that of VGG16 + ConvDet (producing more proposals with less parameters), ranking the second in the published methods. Although MDFN and VGG16 + ConvDet share the same backbone network, reduced VGG-16, MDFN demonstrates prominent advantage. It is noticeable that MDFN performs the best in detecting objects that belong to moderate and hard levels for all the three categories, and the improved performance for the moderate and hard objects contributes to the best final mAP. The AP of pedestrian by MDFN surpasses that of SqueezeDet+ by over 4\%, and its AP of cyclist exceeds that of SqueezeDet+ by over 5\%. Based on above experiments, MDFN, clearly, performs better in the detection of small and occluded objects in cluttered scenes.

\begin{figure*}[th]
	\centering
	\includegraphics[width=0.823\linewidth]{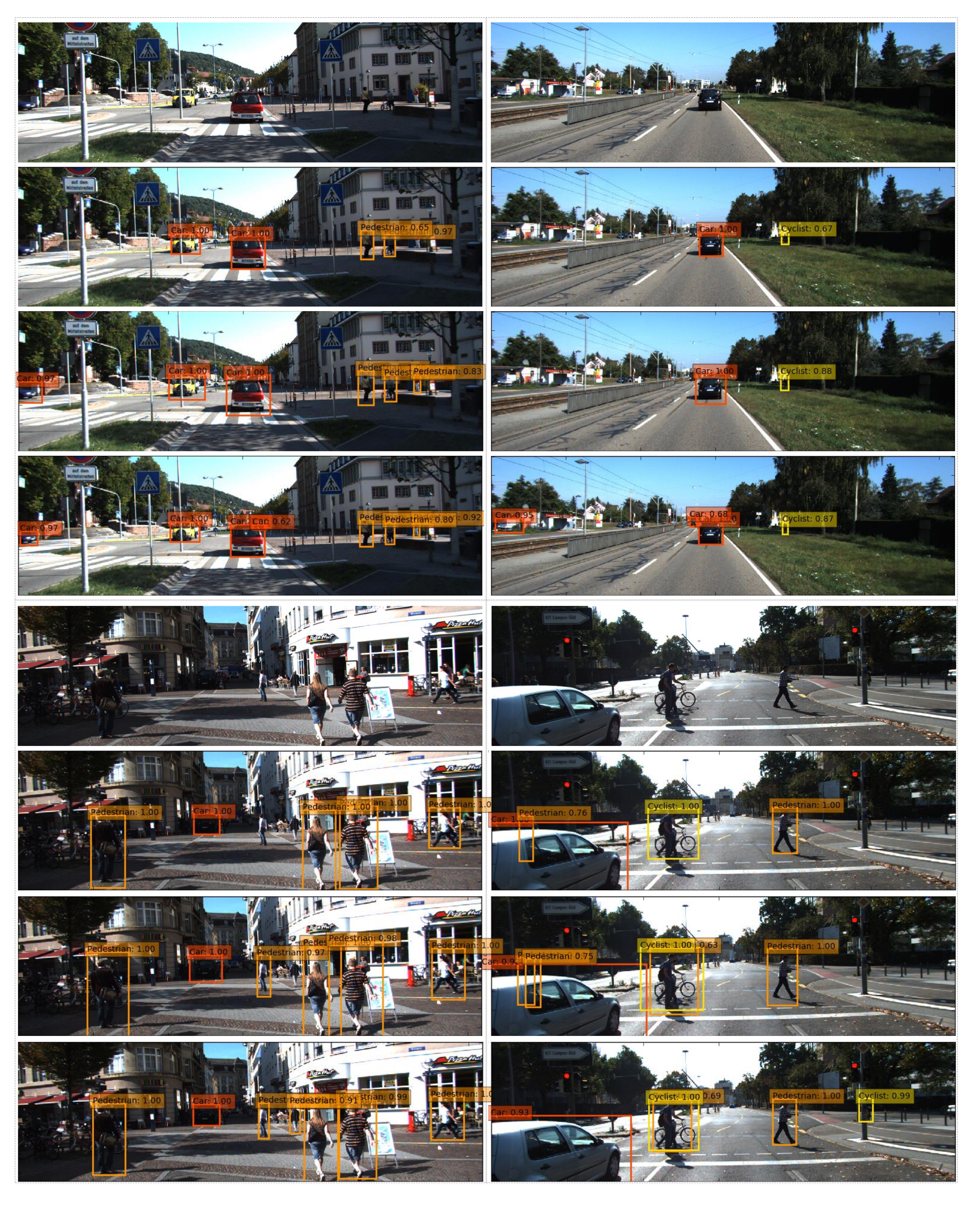}\\%
	%(a)\hspace{0.42\linewidth}(b)\\%[2pt]
	%\vspace{-5mm}
	\caption{\textbf{Four sets of comparative detection examples of SSD and MDFN on KITTI dataset.} In each set, the four images, from top to bottom, represent the original image and the results from SSD, MDFN-I1 and MDFN-I2 respectively. }
	\label{fig:detectionrecog5}
\end{figure*}

Figure~\ref{fig:detectionrecog5} shows several detection examples of SSD, MDFN-I1, and MDFN-I2 on KITTI. There are four sets of images from four different scenes. In each set, the top one is the original image and the other three, from top to bottom, are the results of SSD, MDFN-I1 and MDFN-I2, respectively. In the top-left set, MDFN-I2 detects four cars, three pedestrians with high accuracies, while SSD only detects two cars and two pedestrians. MDFN-I2 is even able to detect the tiny occluded vehicle on the right shoulder of the red car, while all other models miss it. In the top-right set, MDFN-I2 detects two more cars than the other two models, one is in the very left of the image and the other one is the truck in the end of the view. In the bottom-left set, MDFN detects four or five pedestrians, compared to the three of SSD. In the last set, MDFN-I1 detects the pedestrian in parallel with the cyclist and another one behind the car on the left. MDFN-I2 misses the two pedestrians but it is the only model that detects the cyclist occluded by the post on the right of the scene. These four examples show the superior ability of MDFN in detecting small and occluded objects.

\paragraph{Performance under Multiple IoU:} We adopt mAP with different IoU thresholds for further evaluation. In Table~\ref{table:detectionrecog3}, ~\ref{table:detectionrecog4} and ~\ref{table:detectionrecog5}, we provide the performances of SSD and MDFN models with IoU from 0.5 to 0.8 (in steps of 0.05) for Car, Pedestrian and Cyclist, respectively. We can observe that MDFN models perform consistently better than SSD under different IoUs. This experiment further confirms the robustness of the conclusion obtained from Table~\ref{table:detectionrecog2}. It is evident that MDFN-I2 has significant advantage when the IoUs are higher than 0.65 for Cyclist and Pedestrian. For example, for Pedestrian as shown in Table~\ref{table:detectionrecog4}, MDFN-I2 outperforms SSD by 6\% with the IoU of 0.75, and for Cyclist shown in Table~\ref{table:detectionrecog5}, MDFN-I2 outperforms SSD by over 8\% when IoU is 0.8. This experiment further demonstrates that the proposed MDFN models are able to provide more accurate and robust detection performance. %Figure~\ref{fig:detectionrecog4} illustrates the performance curve with respect to different IoU values, where the black and red curves represent MDFN-I1 and MDFN-I2 respectively and the blue one is the results of SSD. It is clear that MDFN models consistently outperform SSD from low IoU to high IoU threshold.

%\begin{figure}[th]
%	\centering
%	\includegraphics[width=0.6\linewidth]{Fig7.pdf}\\%
%	%(a)\hspace{0.42\linewidth}(b)\\%[2pt]
%	\vspace{-1mm}
%	\caption{Mean average precision with respect to IoU.}
%	\label{fig:detectionrecog4}
%\end{figure}

\begin{table}[ht]
	\begin{center}
		\def\arraystretch{0.8}
		\scalebox{0.7}[0.7]{%
			\begin{tabular} {|c|c|c|c|c|c|c|c|c|}
				\hline
				\textbf{Methods} & \textbf{Network} & \textbf{0.5} & \textbf{0.55} & \textbf{0.6} & \textbf{0.65} & \textbf{0.7} & \textbf{0.75} & \textbf{0.8} \\
				\hline
				SSD & VGG & 89.6 & 89.5 & 89.4 & 89.0 & 87.4 & 80.3 & 78.0\\
				\hline
				MDFN-I1 & VGG & 90.0 & 90.0 & 89.8 & \textbf{89.6} & \textbf{88.6} & \textbf{80.4} & 78.5\\
				\hline
				MDFN-I2 & VGG & 90.0 & 90.0 & \textbf{89.9} & 89.5 & 88.5 & 80.3 & 78.5\\
				\hline
			\end{tabular} %
		}
		\def\arraystretch{0.8}	
		\caption{AP of Car on KITTI validation set for different IoU thresholds.}
		\vspace{-5mm}
		\label{table:detectionrecog3}
	\end{center}
\end{table}

\begin{table}[ht]
	\begin{center}
		\def\arraystretch{0.8}
		\scalebox{0.7}[0.7]{%
			\begin{tabular} {|c|c|c|c|c|c|c|c|c|}
				\hline
				\textbf{Methods} & \textbf{Network} & \textbf{0.5} & \textbf{0.55} & \textbf{0.6} & \textbf{0.65} & \textbf{0.7} & \textbf{0.75} & \textbf{0.8} \\
				\hline
				SSD & VGG & 74.8 & 72.4 & 66.1 & 60.8 & 51.8 & 38.2 & 26.7\\
				\hline
				MDFN-I1 & VGG & 76.9 & \textbf{75.3} & 70.2 & 64.5 & 54.8 & 43.4 & 26.4\\
				\hline
				MDFN-I2 & VGG & \textbf{77.2} & 75.2 & 70.2 & \textbf{64.8} & \textbf{55.4} & \textbf{44.1} & 25.9\\
				\hline
			\end{tabular} %
		}
		\def\arraystretch{0.8}	
		\caption{ AP of Pedestrian on KITTI validation set for different IoU thresholds.}
		\vspace{-5mm}
		\label{table:detectionrecog4}
	\end{center}
\end{table}

\begin{table}[ht]
	\begin{center}
		\def\arraystretch{0.8}
		\scalebox{0.7}[0.7]{%
			\begin{tabular} {|c|c|c|c|c|c|c|c|c|}
				\hline
				\textbf{Methods} & \textbf{Network} & \textbf{0.5} & \textbf{0.55} & \textbf{0.6} & \textbf{0.65} & \textbf{0.7} & \textbf{0.75} & \textbf{0.8} \\
				\hline
				SSD & VGG & 82.6 & 81.7 & 77.6 & 76.1 & 68.8 & 61.9 & 46.4\\
				\hline
				MDFN-I1 & VGG & \textbf{86.2} & \textbf{85.8} & \textbf{82.7} & 78.4 & 77.4 & 66.7 & 53.8\\
				\hline
				MDFN-I2 & VGG & 85.8 & 84.7 & 79.1 & \textbf{78.8} & \textbf{77.5} & \textbf{68.0} & \textbf{54.5}\\
				\hline
			\end{tabular} %
		}
		\def\arraystretch{0.8}	
		\caption{ AP of Cyclist on KITTI validation set for different IoU thresholds.}
		\vspace{-5mm}
		\label{table:detectionrecog5}
	\end{center}
\end{table}

\subsection{Detection Results on PASCAL VOC Dataset}

\begin{figure*}[th]
	\centering
	\includegraphics[width=0.95\linewidth]{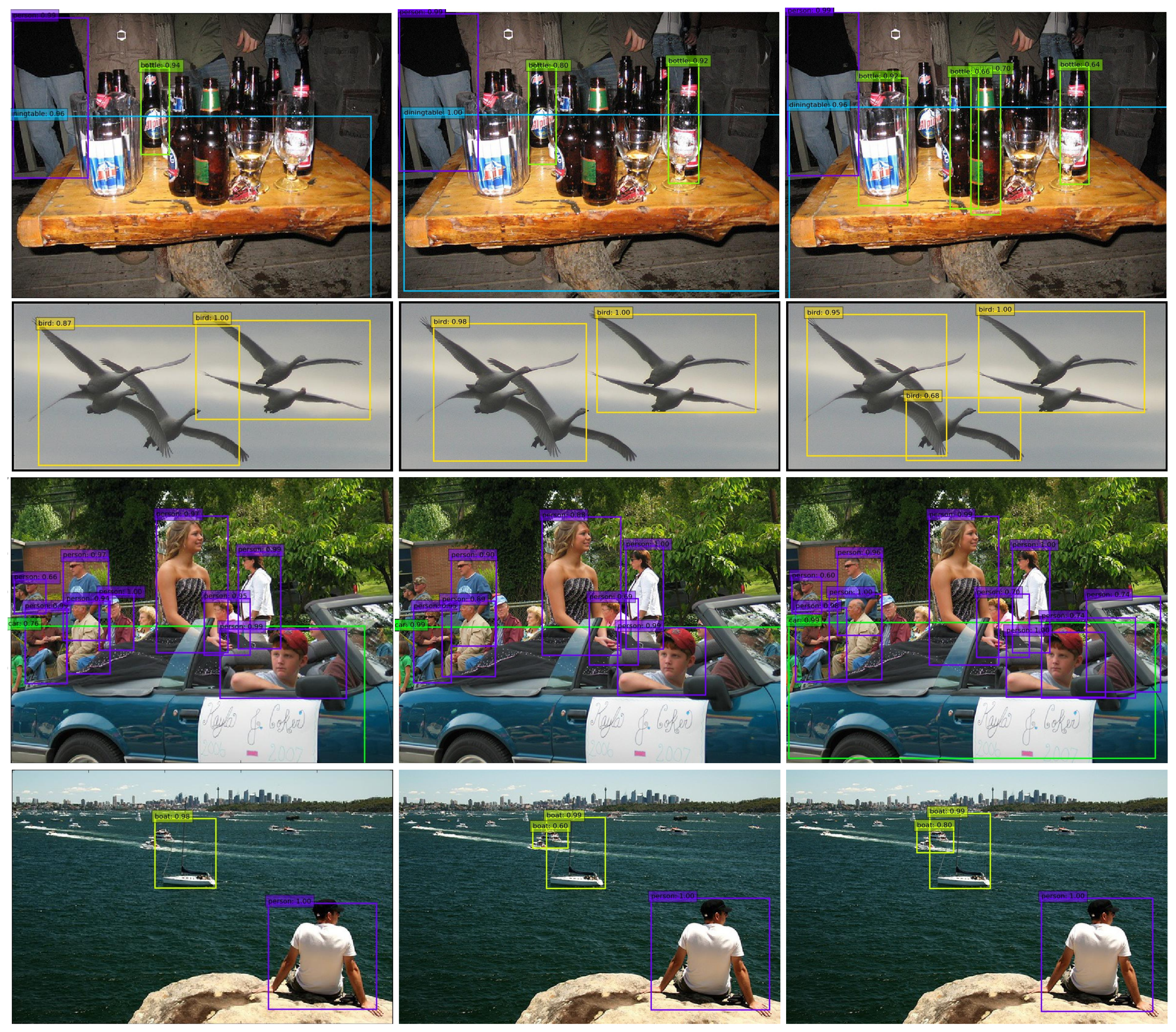}\\%
	%(a)\hspace{0.42\linewidth}(b)\\%[2pt]
	\caption{\textbf{Four sets of comparative detection examples of SSD and MDFN models on VOC2007 dataset.} In each set, the three images, from left to right, represent the results from SSD, MDFN-I1 and MDFN-I2, respectively.}
	\label{fig:detectionrecog6}
\end{figure*}

\begin{table*}[ht]
	\begin{center}
		\def\arraystretch{0.8}
		\scalebox{0.50}[0.50]{%
			\begin{tabular} {|c|c|c|c c c c c c c c c c c c c c c c c c c c|}
				\hline
				\textbf{Model} & \textbf{Network} & \textbf{mAP} & \textbf{aero} & \textbf{bike} & \textbf{bird} & \textbf{boat} & \textbf{bottle} & \textbf{bus} & \textbf{car} & \textbf{cat} & \textbf{chair} & \textbf{cow} & \textbf{table} & \textbf{dog} & \textbf{horse} & \textbf{mbike} & \textbf{person} & \textbf{plant} & \textbf{sheep} & \textbf{sofa} & \textbf{train} & \textbf{tv}\\
				\hline
				Faster I & VGG & 73.2 & 76.5 & 79.0 & 70.9 & 65.5 & 52.1 & 83.1 & 84.7 & 86.4 & 52.0 & 81.9 & 65.7 & 84.8 & 84.6 & 77.5 & 76.7 & 38.8 & 73.6 & 73.9 & 83.0 & 72.6\\
				\hline
				Faster II & Residual-101 & 76.4 & 79.8 & 80.7 & 76.2 & 68.3 & 55.9 & 85.1 & 85.3 & 89.8 & 56.7 & 87.8 & 69.4 & 88.3 & 88.9 & 80.9 & 78.4 & 41.7 & 78.6 & 79.8 & 85.3 & 72.0\\
				\hline
				ION & VGG & 75.6 & 79.2 & 83.1 & 77.6 & 65.6 & 54.9 & 85.4 & 85.1 & 87.0 & 54.4 & 80.6 & 73.8 & 85.3 & 82.2 & 82.2 & 74.4 & 47.1 & 75.8 & 72.7 & 84.2 & 80.4\\
				\hline
				MR-CNN & VGG & 78.2 & 80.3 & 84.1 & 78.5 & 70.8 & 68.5 & 88.0 & 85.9 & 87.8 & 60.3 & 80.52 & 73.7 & 87.2 & 86.5 & 85.0 & 76.4 & 48.5 & 76.3 & 75.5 & 85.0 & 81.0\\
				\hline
				R-FCN & Residual-101 & 80.5 & 79.9 & 87.2 & 81.5 & 72.0 & 69.8 & 86.8 & 88.5 & 89.8 & 67.0 & 88.1 & 74.5 & 89.8 & 90.6 & 79.9 & 81.2 & 53.7 & 81.8 & 81.5 & 85.9 & 79.9\\
				\hline
				YOLOv2 352$\times$352 & Darknet & 73.7 & - & - & - & - & - & - & - & - & - & - & - & - & - & - & - & - & - & - & - & -\\
				\hline
				SSD300$^{*}$ & VGG & 77.5 & 79.5 & 83.9 & 76.0 & 69.6 & 50.5 & 87.0 & 85.7 & 88.1 & 60.3 & 81.5 & 77.0 & 86.1 & 87.5 & 83.97 & 79.4 & 52.3 & 77.9 & 79.5 & 87.6 & 76.8\\
				\hline
				SSD512$^{*}$ & VGG & 79.5 & 84.8 & 85.1 & 81.5 & 73.0 & 57.8 & 87.8 & 88.3 & 87.4 & 63.5 & 85.4 & 73.2 & 86.2 & 86.7 & 83.9 & 82.5 & 55.6 & 81.7 & 79.0 & 86.6 & 80.0\\
				\hline
				SSD300 & VGG & 72.1 & 75.2 & 79.8 & 70.5 & 62.5 & 41.3 & 81.1 & 80.8 & 86.4 & 51.5 & 74.3 & 72.3 & 83.5 & 84.6 & 80.6 & 74.5 & 46.0 & 71.4 & 73.8 & 83.0 & 69.1 \\
				\hline
				SSD500 & VGG & 75.1 & 79.8 & 79.5 & 74.5 & 63.4 & 51.9 & 84.9 & 85.6 & 87.2 & 56.6 & 80.1 & 70.0 & 85.4 & 84.9 & 80.9 & 78.2 & 49.0 & 78.4 & 72.4 & 84.6 & 75.5 \\
				\hline
				SSD321 & Residual-101 & 74.8 & 76.0 & 84.9 & 74.6 & 62.4 & 44.8 & 84.9 & 82.9 & 86.2 & 57.6 & 79.9 & 71.2 & 86.2 & 87.4 & 83.4 & 77.0 & 45.5 & 74.1 & 75.9 & 86.1 & 75.4 \\
				\hline
				MDFN-I1-321 (ours) & Residual-101 & 75.9 & 76.8 & 83.5 & 74.7 & 65.8 & 46.5 & 85.2 & 83.7 & 88.2 & 59.9 & 78.3 & 74.3 & 86.7 & 87.4 & 83.8 & 78.1 & 47.4 & 76.1 & 81.0 & 86.2 & 74.3\\
				\hline					
				MDFN-I2-321 (ours) & Residual-101 & 77.0 & 78.0 & 86.0 & 78.0 & 67.5 & 49.4 & 86.2 & 83.8 & 87.6 & 59.7 & 80.8 & 76.6 & 86.8 & 87.5 & 85.0 & 78.5 & 49.6 & 75.5 & 80.3 & 86.3 & 76.3 \\
				\hline								
				YOLOv2 544$\times$544 & Darknet & 78.6 & - & - & - & - & - & - & - & - & - & - & - & - & - & - & - & - & - & - & - & -\\
				\hline
				CC-Net & CC-Net & 80.4 & 83.0 & 85.8 & 80.0 & 73.4 & 64.6 & 88.3 & 88.3 & 89.2 & 63.2 & 86.0 & 76.8 & 87.6 & 88.2 & 83.4 & 84.1 & 54.9 & 83.7 & 77.7 & 86.0 & 83.6 \\
				\hline
				BlitzNet512(s8) & ResNet-50 & 80.7 & 87.7 & 85.4 & 83.6 & 73.3 & 58.5 & 86.6 & 87.9 & 88.5 & 63.7 & 87.3 & 77.6 & 87.3 & 88.1 & 86.2 & 81.3 & 57.1 & 84.9 & 79.8 & 87.9 & 81.5 \\
				\hline
				MDFN-I1 (ours) & VGG & 79.3 & 81.2 & 87.0 & 79.2 & 72.3 & 57.0 & 87.3 & 87.1 & 87.5 & 63.1 & 84.2 & 76.7 & 87.6 & 88.8 & 85.6 & 81.0 & 56.0 & 80.4 & 79.9 & 88.0 & 77.0 \\
				\hline
				MDFN-I2 (ours) & VGG & 78.3 & 82.5 & 85.9 & 78.0 & 70.5 & 54.0& 87.9 & 87.4 & 88.6 & 60.3 & 82.6 & 73.7 & 86.7 & 87.5 & 85.0 & 80.6 & 52.3 & 77.9 & 80.6 & 88.1 & 76.6\\
				\hline
			\end{tabular} %
		}
		\def\arraystretch{0.8}	
		\caption{PASCAL VOC2007 test detection results.}
		\vspace{-3mm} 
		\label{table:detectionrecog6}
	\end{center}
\end{table*}

Table~\ref{table:detectionrecog6} shows the detection AP on PASCAL VOC 2007 test benchmark. The leaderboard provides current state-of-the-art detection results, including Faster I~\cite{ren2015faster}, Faster II~\cite{he2016deep}, ION~\cite{bell2016inside}, MR-CNN~\cite{gidaris2015object}, R-FCN~\cite{dai2016r}, YOLOv2 352$\times$352~\cite{redmon2017yolo9000}, SSD300$^{*}$~\cite{liu2016ssd}, SSD 321~\cite{fu2017dssd}, SSD512$^{*}$~\cite{liu2016ssd}, SSD300~\cite{liu2016ssd}, SSD500~\cite{liu2016ssd}, YOLOv2 544$\times$544~\cite{redmon2017yolo9000}, CC-Net~\cite{ouyang2017chained} and BlitzNet512(s8)~\cite{dvornik2017blitznet}. In this leading board, MDFN models obtain leading performance in terms of mAP and achieve top AP in the object class of train. MDFN models are the only ones that exceed 88\% in the detection of trains. MDFN-I1 ranks the second in detecting plant. The proposed models also achieve very competing performance, though not the best, in other categories. Please note that some methods that outperform ours adopt the very deep ResNet as their base network. For example, R-FCN obtained mAP over 80\% by introducing Residual-101 as their base network, and BlitzNet512 reaches 80.7\% by adopting ResNet-50 as its backbone network. 

As discussed in Section 1, introducing much deeper ResNet as base network may result in higher detection accuracy which can be noted by the results shown in Table~\ref{table:detectionrecog6} where SSD321 with ResNet-101 obtains higher mAP than SSD300 with VGG, and the performance of the former is close to that of SSD500 with VGG. On one hand, the proposed MDFN models obtain competitive detection accuracy with VGG compared to the models with ResNet-101 like BlitzNet512~\cite{dvornik2017blitznet} which jointly performs object detection and segmentation in one forward pass for better scene understanding; on the other hand, the detector with ResNet-101 as base network would prominently reduce the running speed of the model. As shown in Table~\ref{table:detectionrecog7}, the models with ResNet-101 generally perform slower than other models with VGG. Specifically, SSD300 with VGG-16 runs at 84 frames per second, more than two times faster than that of SSD321 with ResNet-101. MDFN model with VGG-16 on the image scale of 500 runs even faster than SSD321 with ResNet-101. In Table~\ref{table:detectionrecog6}, the symbol $*$ in SSD300$^{*}$ and SSD512$^{*}$ indicate these two methods introduce a data augmentation trick called random expansion, while MDFN does not adopt it. %Please note that the trick has no influence on the test time.

Figure~\ref{fig:detectionrecog6} shows some comparative detection examples on VOC, where MDFN models perform better in complicated scenes. For example, MDFN-I2 detects four bottles while SSD only two in the first set. In the second set, one more bird is detected by MDFN-I2. The smaller boat in the fourth set is detected by MDFN models while SSD misses it. In the last set, more persons especially the two men occluded by the front boy in red hat are detected by MDFN-I2. While compared by mAP, MDFN-I2 is not higher than MDFN-I1. More discussion and analysis towards this phenomenon will be given in the later subsection.

\subsection{Detection Results on COCO}

The detection AP on COCO benchmark is shown in Table~\ref{table:detectionrecog_coco}, where comparative results of mainstream detection models are provided as well, including Faster I~\cite{ren2015faster, liu2016ssd}, Faster II~\cite{liu2016ssd}, YOLOv2~\cite{redmon2017yolo9000}, YOLOv3~\cite{redmon2018yolov3}, SSD300~\cite{liu2016ssd}, and SSD512~\cite{liu2016ssd}. The proposed MDFN-I2 model on the image resolution of 512 leads the board with the highest accuracy under the three criteria of AP, AP$_{50}$, and AP$_{75}$, where AP refers to the average precision over 10 IoU levels on 80 categories (AP@[.50:.05:.95]: start from 0.5 to 0.95 with a step size of 0.05). We can observe that MDFN models show clear advantage on all the three criteria over both two-stage detectors (\ie Faster I and Faster II), and one-stage detectors (\ie YOLO and SSD). Especially for AP$_{75}$, MDFN shows the advantage with MDFN-I2-512 yielding over 30\%. MDFN performs better for hard objects, in accord with that on KITTI and VOC datasets. In contrast, although both YOLOv2 and YOLOv3 achieve higher precision for AP$_{50}$, the precision of AP and AP$_{75}$ reveal that they perform inferior on hard situations. This reflects the weakness of YOLO models that they struggle in recognizing the accurate locations of the objects. It is evident that, on both image scales of 300 and 512, MDFN yields robust better performance in comparison with SSD, which supports the viewpoint that deep features are semantically abstract and suitable for extracting global visual primitives. The better performance of MDFN-I2 over MDFN-I1 is due to the deeper multi-scale feature learning, which is more obvious for higher IoU threshold, like the significant 4.6\% improvement from SSD300 to MDFN-I2-300 at the IoU of 0.75, compared with the 2.2\% at 0.5. For image scale of 500, MDFN demonstrates the same advantage. These results are consistent with the theoretical analysis in Section \uppercase\expandafter{\romannumeral3} and match the performance on the other two benchmarks. 

\begin{table}[ht]
	\begin{center}
		\def\arraystretch{1.0}
		\scalebox{0.7}[0.7]{%
			\begin{tabular} {|c|c|c|c|c|c|}
				\hline
				\textbf{Method} & \textbf{Data} & \textbf{Backbone} & \textbf{AP} & \textbf{AP$_{50}$} & \textbf{AP$_{75}$} \\
				\hline
				Faster I & trainval & VGG & 21.9 & 42.7 & -\\
				\hline
				Faster II & trainval & ResNet-101 & 24.2 & 45.3 & 23.5 \\
				\hline
				YOLOv2 & trainval35k & DarkNet-19 & 21.6 & 44.0 & 19.2 \\
				\hline
				SSD300 & trainval35k & VGG & 23.2 & 41.2 & 23.4 \\
				\hline
				MDFN-I1-300 (ours) & trainval35k & VGG & 26.3 & 42.9 & 26.9 \\
				\hline
				MDFN-I2-300 (ours) & trianval35k & VGG & 27.1 & 43.4 & 28.0 \\
				\hline
                YOLOv3-320 & trainval35k & DarkNet-53 & 28.2 & 51.5 & - \\
                \hline		
				SSD512 & trainval35k & VGG & 26.8 & 46.5 & 27.8 \\
				\hline
				MDFN-I1-512 (ours) & trainval35k & VGG & 28.9 & 47.6 & 29.9\\
				\hline
				MDFN-I2-512 (ours) & trainval35k & VGG & 29.6 & 48.1 & 30.8 \\
				\hline
			\end{tabular} %
		}
		\def\arraystretch{1.0}	
		\caption{Detection results on COCO test-dev}
		\vspace{-3mm}
		\label{table:detectionrecog_coco}
	\end{center}
\end{table}

\subsection{Multi-scale Feature Depth vs. Performance}

In this study, we have proposed two models, MDFN-I1 and MDFN-I2, with the purpose of exploring the effectiveness of features produced at different network depths. Here network depth is defined as the maximum layer depth that deep multi-scale features are extracted from, among the deep layers after the base network. For MDFN-I1, its multi-scale feature depth is 3, and it is 4 for MDFN-I2. MDFN-I2 extracts the deep features from all the four deep layers before the prediction layer. Theoretically, this would enhance the ability of feature expression and scene understanding. However, according to the mAP results on both KITTI and VOC2007, MDFN-I2 does not surpass MDFN-I1. While if we increase the threshold of IoU, MDFN-I2 shows its advantage of higher accuracy, which can be evidenced especially from Table~\ref{table:detectionrecog4} and Table~\ref{table:detectionrecog5}. 

\begin{figure*}[th]
	\centering
	\includegraphics[width=0.9\linewidth]{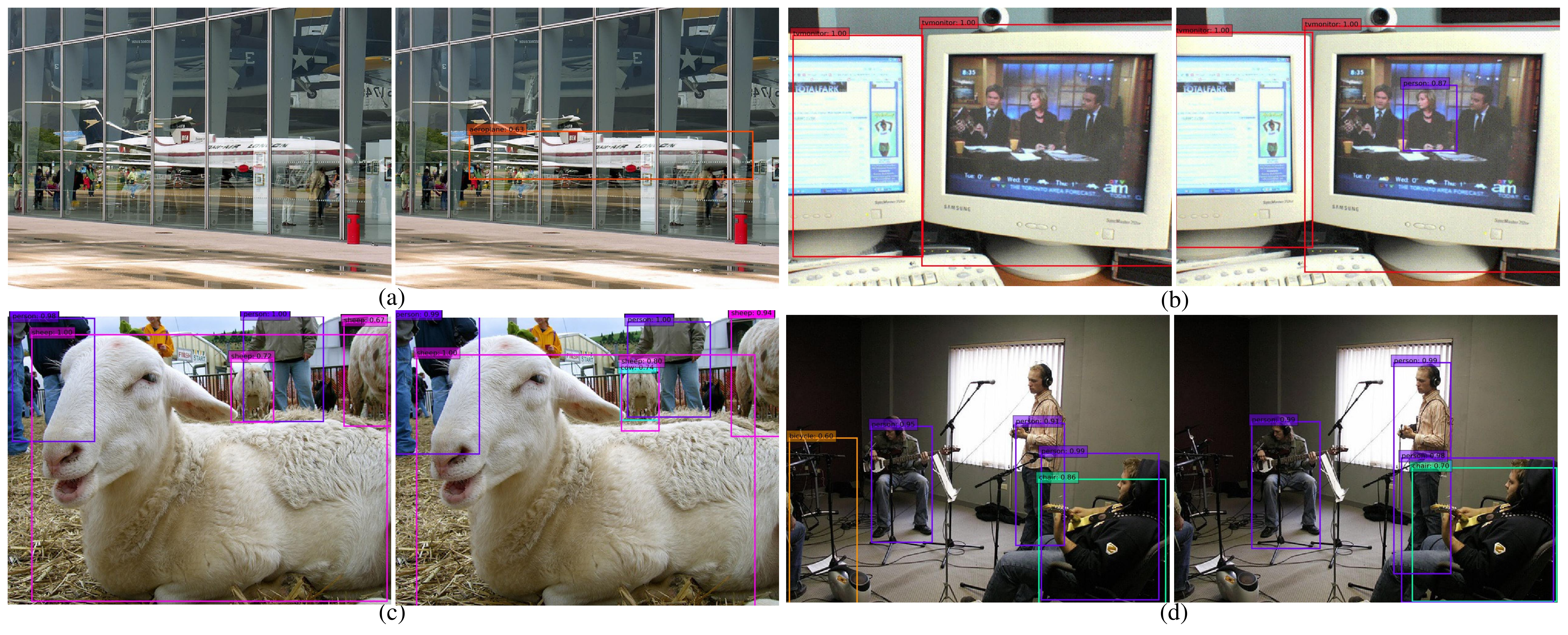}\\%
	%(a)\hspace{0.42\linewidth}(b)\\%[2pt]
	\caption{\textbf{Comparative detection examples by the two MDFN models.} In each set, the left and right images refer to the detection results of MDFN-I1 and MDFN-I2 respectively.}
	\label{fig:detectionrecog7}
\end{figure*}

From Figure~\ref{fig:detectionrecog7} (a) and (b), MDFN-I2 is capable of detecting the reflected objects, like the airplane in (a) and the lady on the screen in (b). These objects usually have fuzzy features or much smaller sizes than true objects. Thus, the successful detection of them exhibits the power of MDFN-I2. This is one reason that would lower the mAP. From Figure~\ref{fig:detectionrecog7} (c), MDFN-I2 yields multiple choices for some uncertain objects like the sheep back to the view, which is detected as a sheep or cow. The mistaken detection of objects with limited features also leads to the decrease of mAP. Nevertheless, MDFN-I2, by considering more context information, brings in more accurate localization and better classification in most situations like the bandsman in (d), where MDFN-I2 locates his figure completely compared to the part recognition given by MDFN-I1. Considering the space, we only show limited examples here. But MDFN-I2 occasionally fails to detect the partially occluded person, like the man behind the flowering shrubs, whose lower half is occluded, or a lady only showing her head in a group photo. This leak detection is possible as MDFN-I2 that considers the context information more than MDFN-I1, especially when this kind of objects account a tiny portion of the entire dataset. This can be another reason that MDFN-I2 does not beat MDFN-I1 in overall mAP on PASCAL VOC.

\subsection{Efficiency Discussion}

\begin{table}[ht]
	\begin{center}
		\def\arraystretch{1.0}
		\scalebox{0.7}[0.7]{%
			\begin{tabular} {|c|c|c|c|c|c|c|c|}
				\hline
				\textbf{Method} & \textbf{Data} & \textbf{Resolution} & \textbf{network} & \textbf{\#Params} & \textbf{GPU} & \textbf{FPS} & \textbf{FLOPS} \\
				\hline
				SSD & KITTI & 1242$\times$375 & VGG-16 & 24.0M & Titan Xp & 30 & 157.4G\\
				\hline
				MDFN-I1 & KITTI & 1242$\times$375 & VGG-16 & 26.8M & Titan Xp & 28 & 158.2G\\
				\hline
				MDFN-I2 & KITTI & 1242$\times$375 & VGG-16 & 27.0M & Titan Xp & 27 & 158.3G\\
				\hline
				Faster I & VOC2007 & ~1000$\times$600 & VGG-16 & 144.8M & Titan X & 7 & 184.0G\\
				\hline
				Faster II & VOC2007 & ~1000$\times$600 & Residual-101 & 734.7M & K40 & 2.4 & 93.0G\\
				\hline
				R-FCN & VOC2007 & ~1000$\times$600 & Residual-101 & - & Titan X & 9 & -\\
				\hline
				YOLOv2 & VOC2007 & 352$\times$352 & Darknet-19 & 48.6M & Titan X & 81 & 9.06G\\
				\hline
				YOLOv2 & VOC2007 & 544$\times$544 & Darknet-19 & 48.6M & Titan X & 40 & 21.65G\\
				\hline
				SSD300$^{*}$ & VOC2007 & 300$\times$300 & VGG-16 & - & Titan X & 46 & -\\
				\hline
				SSD512$^{*}$ & VOC2007 & 512$\times$512 & VGG-16 & - & Titan X & 19 & -\\
				\hline
				SSD300 & VOC2007 & 300$\times$300 & VGG-16 & 26.3M & Titan Xp & 84 & 31.4G\\
				\hline				
				SSD321 & VOC2007 & 321$\times$321 & Residual-101 & 52.7M & Titan Xp & 30 & 22.1G\\
				\hline
				MDFN-I1-321 & VOC2007 & 321$\times$321 & Residual-101 & 61.6M & Titan Xp & 26 & 22.55G \\
				\hline				
				MDFN-I2-321 & VOC2007 & 321$\times$321 & Residual-101 & 64.2M & Titan Xp & 26 & 22.56G\\
				\hline
				%SSD513 & VOC2007 & 513$\times$513 & Residual-101 & 41.1M & Titan X & 6.8 & 56.6G\\
				%\hline
				SSD & VOC2007 & 500$\times$500 & VGG-16 & 26.3M & Titan Xp & 45 & 87.2G\\
				\hline
				MDFN-I1 & VOC2007 & 500$\times$500 & VGG-16 & 30.8M & Titan Xp & 39 & 87.9G\\
				\hline
				MDFN-I2 & VOC2007 & 500$\times$500 & VGG-16 & 31.1M & Titan Xp & 38 & 88.0G\\
				\hline
				SSD & COCO & 300$\times$300 & VGG-16 & 34.3M & Titan Xp & 75 & 34.4G\\
				\hline
				MDFN-I1 & COCO & 300$\times$300 & VGG-16 & 44.6M & Titan Xp & 60 & 35.1G\\
				\hline
				MDFN-I2 & COCO & 300$\times$300 & VGG-16 & 45.6M & Titan Xp & 58 & 35.1G\\
				\hline
				YOLOv3 & COCO & 320$\times$320 & DarkNet-53 & 61.9M & Titan X & 45 & 16.14G\\
				\hline
				SSD & COCO & 500$\times$500 & VGG-16 & 34.3M & Titan Xp & 41 & 98.7G\\
				\hline
				MDFN-I1 & COCO & 500$\times$500 & VGG-16 & 44.6M & Titan Xp & 35 & 100.4G\\
				\hline
				MDFN-I2 & COCO & 500$\times$500 & VGG-16 & 45.6M & Titan Xp & 35 & 100.6G\\
				\hline																
			\end{tabular} %
		}
		\def\arraystretch{1.0}	
		\caption{Comparison of inference time on KITTI, VOC2007 and COCO test datasets.} 
		\label{table:detectionrecog7}
	\end{center}
\end{table}

\begin{figure}[h]
	\centering
	\includegraphics[width=0.6\linewidth]{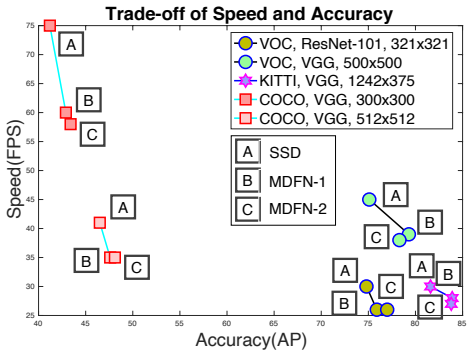}\\%
	%(a)\hspace{0.42\linewidth}(b)\\%[2pt]
	\vspace{-5mm}
	\caption{The trade-off between the accuracy and speed.}
	\vspace{-2mm}
	\label{fig:detectionrecog8}
\end{figure}

In Table~\ref{table:detectionrecog7}, we show a comparison of the inference time on KITTI, VOC2007, and COCO benchmarks. The compared frameworks include SSD~\cite{liu2016ssd}, Faster I~\cite{ren2015faster} with VGG-16, Faster II~\cite{he2016deep} with Residual-101, R-FCN~\cite{dai2016r}, YOLOv2~\cite{redmon2017yolo9000}, YOLOv3~\cite{redmon2018yolov3}, SSD300$^{*}$~\cite{liu2016ssd}, SSD512$^{*}$~\cite{liu2016ssd}, and SSD321~\cite{fu2017dssd}. Compared with the one-stage detectors, two-stage detectors (\ie Faster I and Faster II) normally run much slower (less ``FPS'') and require more parameters (larger ``\#Params''). As shown in Table \ref{table:detectionrecog7}, %in running speed represented by \hilight{much less ``FPS'' and in model size by much larger ``\#Params''}, where 
the parameter number of Faster II is over 10 times of that of MDFN. Among the one-stage detectors, YOLO runs the fastest in general. For example, YOLOv2 performs at 81 FPS, which is more than one time of that of SSD300$^{*}$ (\ie 46 FPS). This is due to the design of YOLO by one end-to-end inference process and limited objects' proposals. However, it can be seen that the YOLO models are generally built by more parameters than SSD and MDFN. For example, the ``\#Params'' of YOLOv3, is nearly two times of those of VGG16-based SSD and MDFN. Thus, when we consider the efficiency of models, YOLO can barely be regarded as highly-effective since more parameters usually mean more demand of memory in real-life applications. 

On the contrary, MDFN makes a better trade-off between the detection accuracy and operational efficiency. Specifically, although the introduction of deep feature learning modules brings around 10\% increase of the number of parameters, the running speed of the MDFN models only decrease less than 4\% and the flops only increase around 2\%, shown on KITTI and COCO. As shown in Figure \ref{fig:detectionrecog8}, notations A, B and C form the similar pattern like a triangle on each benchmark and B, C are not far from A along the speed axis, indicating the robustness and stability of MDFN and the prominent speed and accuracy trade-off. This supports our claim in Section \uppercase\expandafter{\romannumeral3} that the processing towards deep features contributes less computational load compared with the processing for features from earlier layers. Therefore, enhancing the learning ability towards deep features is definitely a high-productive choice. 

%The proposed information square and cubic inception modules certainly promote this advantage by parameter sharing. From Table~\ref{table:detectionrecog6} and Table~\ref{table:detectionrecog7}, the MDFN models significantly outperform those which do not adopt Residual-101 on mAP, and is very closed to those with Residual-101. The results demonstrate that, by introducing deep feature learning, MDFN has achieved a better balance between the higher detection accuracy and more efficient operational capability.

\section{Conclusion}

In this paper, we have proposed a novel multi-scale deep feature learning convolution neural network (MDFN) for object detection. We make full use of the deep features which provide abundant semantic and contextual information by integrating the proposed information square and cubic inception modules into the deep layers of the network. Extensive experiments show that MDFN achieves more accurate localization and classification results on general object detection (VOC), autonomous driving task (KITTI) and full scene understanding (COCO), resulting in consistent and robust semantic representation. The relevant context information from the deep features plays a vital role in detecting hard objects like small and occluded ones. To our best knowledge, MDFN is the first single-shot object detector that specifically focuses on deep feature learning. The proposed approach advances the state-of-the-art techniques in object detection and classification.

\section*{Acknowledgment}
The work was supported in part by USDA NIFA under the award no. 2019-67021-28996, KU General Research Fund (GRF), and the Nvidia GPU grant.
%\section{Front matter}

%The author names and affiliations could be formatted in two ways:
%\begin{enumerate}[(1)]
%\item Group the authors per affiliation.
%\item Use footnotes to indicate the affiliations.
%\end{enumerate}
%See the front matter of this document for examples. You are recommended to conform your choice to the journal you are submitting to.

%\section{Bibliography styles}

%There are various bibliography styles available. You can select the style of your choice in the preamble of this document. These styles are Elsevier styles based on standard styles like Harvard and Vancouver. Please use Bib\TeX\ to generate your bibliography and include DOIs whenever available.

%Here are two sample references: \cite{Feynman1963118,Dirac1953888,wang2018pedestrian}.

%\newpage
%\section*{References}

%\bibliography{mybibfile}
{\small
	\bibliographystyle{elsarticle-num}
	\bibliography{mybibfile}
}

\end{document}